\documentclass[10pt,twocolumn,letterpaper]{article}

\usepackage[pagenumbers]{cvpr} 

\usepackage{multirow}
\usepackage[table]{xcolor}
\usepackage{multicol}
\usepackage{colortbl}
\usepackage{multirow}
\definecolor{dark-gray}{gray}{0.30}
\newcommand{\pub}[1]{{\color{dark-gray}{\scriptsize{[{#1}]}}}}

\usepackage[T1]{fontenc}

\definecolor{cvprblue}{rgb}{0.21,0.49,0.74}
\usepackage[pagebackref,breaklinks,colorlinks,allcolors=cvprblue]{hyperref}

\def\paperID{xxx}
\def\confName{CVPR}
\def\confYear{2026}

\title{MindDriver: Introducing Progressive Multimodal Reasoning \\ for Autonomous Driving}

\author{
   \textbf{Lingjun Zhang$^{1}$\thanks{Equal contribution with random order. Each co-first author may list themselves as lead author on their CV.}\kern0.3em\thanks{Work done during internship at Amap, Alibaba Group.},  
    Yujian Yuan$^{1,2}$\footnotemark[1]\kern0.3em\footnotemark[2],  
    Changjie Wu$^{1}$\thanks{Corresponding author and project leader.},  
    Xinyuan Chang$^{1}$,
    Xin Cai$^{3}$}, \\
    \textbf{Shuang Zeng$^{1,4}$\footnotemark[2],
    Linzhe Shi$^{1}$,
    Sijin Wang$^{1}$,
    Hang Zhang$^{1}$,
    Mu Xu$^{1}$
    }\\
    $^{1}$ Amap, Alibaba Group, $^{2}$ The Hong Kong University of Science and Technology\\
    $^{3}$ The Chinese University of Hong Kong
    $^{4}$ Xi'an Jiaotong University
    \\
    {\tt\small 
    zhanglingjun.zlj@alibaba-inc.com,yyuanbn@connect.ust.hk}
    \\
    {\tt\small 
    \{wuchangjie.wcj,changxinyuan.cxy,sijin.wsj,suishou.zh,xumu.xm\}@alibaba-inc.com}
}

\usepackage{float}
\usepackage{array}   
\usepackage{booktabs}
\begin{document}
\maketitle
\begin{abstract}

Vision-Language Models (VLM) exhibit strong reasoning capabilities, showing promise for end-to-end autonomous driving systems. Chain-of-Thought (CoT), as VLM's widely used reasoning strategy, is facing critical challenges. Existing textual CoT has a large gap between text semantic space and trajectory physical space. Although the recent approach utilizes future image to replace text as CoT process, it lacks clear planning-oriented objective guidance to generate images with accurate scene evolution. To address these, we innovatively propose \textbf{MindDriver}, a \textbf{progressive multimodal reasoning} framework that enables VLM to imitate human-like progressive thinking for autonomous driving.
MindDriver presents semantic understanding, semantic-to-physical space imagination, and physical-space trajectory planning.
To achieve aligned reasoning processes in MindDriver, we develop a \textbf{feedback-guided automatic data annotation pipeline} to generate aligned multimodal reasoning training data. 
Furthermore, we develop a \textbf{progressive reinforcement fine-tuning}
method to optimize the alignment through progressive high-
level reward-based learning.
MindDriver demonstrates superior performance in both nuScences open-loop and Bench2Drive closed-loop evaluation.
Codes are available at: \url{https://github.com/hotdogcheesewhite/MindDriver}.

\end{abstract}    
\section{Introduction}
\label{sec:intro}
Recently, Multimodal Large Language Models (MLLMs) have gained substantial attention in the field of end-to-end autonomous driving~\cite{wang2024omnidrive,zeng2025futuresightdrive,RDA-Driver}.
Their popularity arises from the ability to harness extensive world knowledge obtained through large-scale pre-training, and to align multimodal representations effectively.
One notable approach is utilizing pretrained vision-language models (VLM) to directly extract high-level semantic features from raw sensor inputs and predict vehicle trajectories in physical space~\cite{zeng2025futuresightdrive}. 
This efficient end-to-end architecture simplifies the overall framework, reduces information loss, and leverages world knowledge to understand driving scenes and ensure safe planning in challenging and long-tail scenarios.

\begin{figure}[t]
    \centering
    \includegraphics[width=0.47\textwidth]{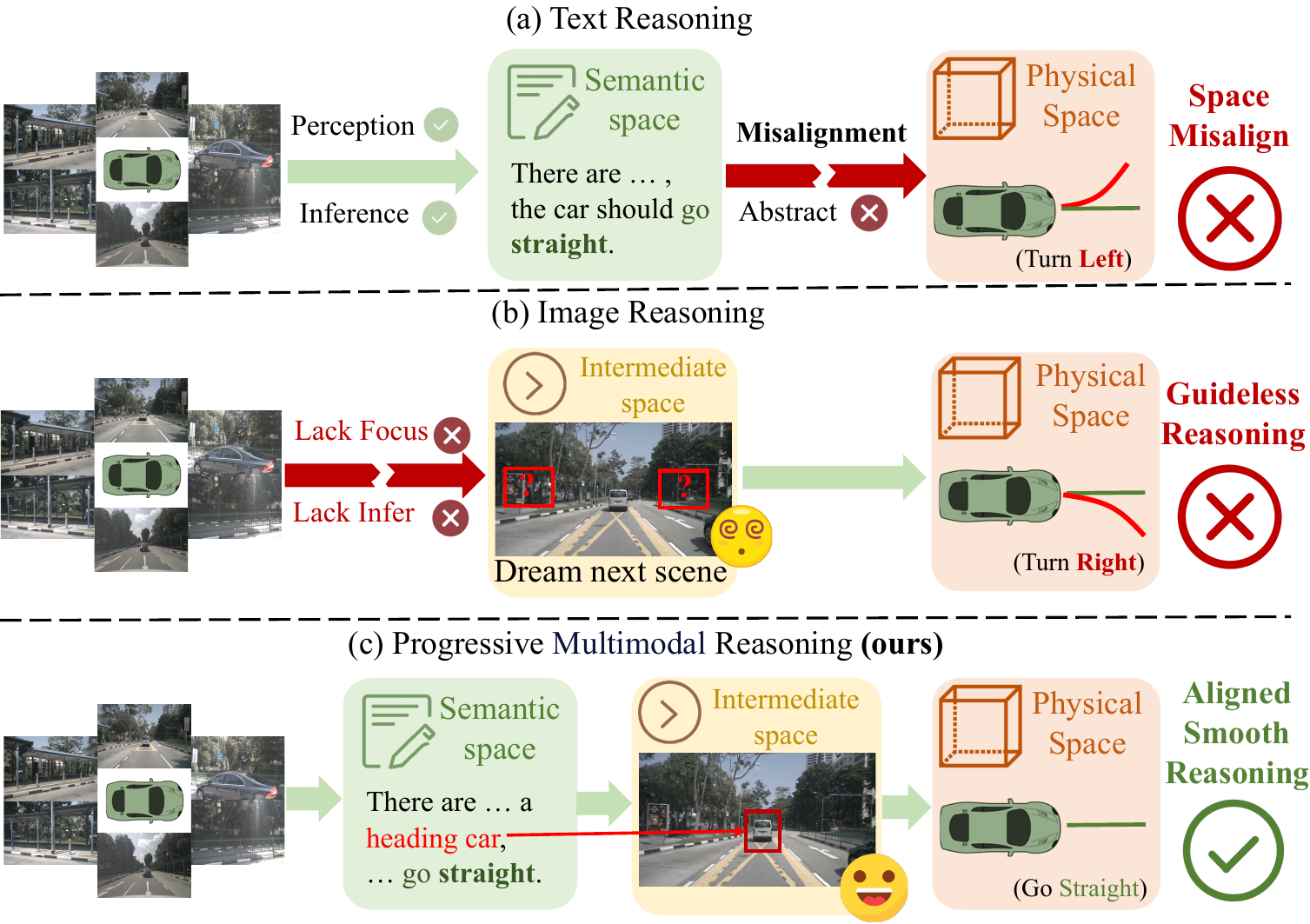}
    \caption{Comparison of different reasoning methods. 
    Text reasoning struggles with space misalignment, while image reasoning suffers from guideless image prediction. 
    Our proposed progressive multimodal reasoning conducts aligned smooth reasoning.}
    \label{fig:fig1}
\end{figure}

Chain-of-Thought (CoT) reasoning, a widely adopted reasoning strategy in VLMs~\cite{Qwen2.5-VL,Qwen2-VL}, has recently been applied to autonomous driving to enhance scene reasoning capabilities and improve the interpretability of driving decisions~\cite{zhou2025autovla}.
As shown in Fig.~\ref{fig:fig1},
benefiting from the large-scale pre-training of LLMs,
traditional CoT methods mainly focus on text reasoning within the semantic space, 
abstractly analyzing the scene and the driving logic.
However, trajectory planning in autonomous driving depends on predictions in the physical space.
Traditional text reasoning that directly predicts trajectory after textual reasoning faces a significant space misalignment between the semantic and physical space, thus resulting in decision misalignment. 
To alleviate this misalignment, recent research has explored using images instead of text as the intermediary for reasoning~\cite{zeng2025futuresightdrive}. 
Images inherently combine semantic and physical information as the intermediate space, which is more suitable to assist trajectory prediction.
However, purely image-driven reasoning lacks a clear planning-oriented objective guidance, which confuses the model about which objects to focus on.
Additionally, image reasoning fails to effectively utilize the extensive driving knowledge embedded in large-scale pretraining of LLMs, thereby limiting its performance in complex and long-tail driving scenarios.

To address the above challenges, 
inspired by the human perception-imagination-action mechanism, we propose a \textbf{Progressive Multimodal Reasoning} method that enables smooth reasoning from textual semantics, through intermediate imagined scene images, to physical trajectories.
As shown in Fig.~\ref{fig:fig1},
our method comprises three key components: (1) Semantic understanding: derive high-level driving insights through textual reasoning for scene understanding, logical decision-making, and so on; (2) Visual imagination: leverage text reasoning as guidance to generate future scene images, bridging semantic and physical spaces.
 and (3) Physical trajectory prediction: leverage the dreamed scene image to predict physically-grounded trajectories.
  This end-to-end thinking ensures progressive and smooth reasoning, enabling effective and interpretable autonomous driving planning.

To achieve progressive multimodal reasoning, we propose MindDriver, 
a novel framework with smooth reasoning from semantic understanding, through semantic-to-physical-space imagination, to physical-space trajectory planning.
However, there are significant challenges when training such a well-aligned multi-component reasoning model:
(1) Lack of Training Dataset: high-quality aligned progressive multimodal reasoning training data is in demand.  (2) Inefficient Training Strategy: traditional supervised fine-tuning focuses on token-level supervision, neglecting the alignment of each component.
To address these, we propose a feedback-guided data annotation framework that automatically aligns each component in reasoning process, by three filtering and feedback-guided re-annotation.
Furthermore, to effectively train the progressive reasoning with well-alignment, we propose a progressive reinforcement fine-tuning post-training method. 
It structures the training process into progressive steps, prioritizing gradual alignment for transitions from semantic understanding to visual imagination, and from visual imagination to trajectory planning.

Extensive experiments on both open-loop~\cite{caesar2020nuscenes} and closed-loop~\cite{jia2024bench2drive} trajectory planning, future frames generation demonstrate the effectiveness of progressive multimodal reasoning, auto-annotation pipeline, and reinforcement fine-tuning in MindDriver.

In summary, our contributions include:

\begin{itemize}
    \item We propose a progressive multimodal reasoning method that enhances model's trajectory planning by smooth reasoning from text semantic understanding, through intermediate imagined scene images, to physical trajectories.

    \item To ensure reasoning alignment in MindDriver, we design a feedback-guided automatic data annotation framework to generate aligned multimodal reasoning data. Furthermore, we develop a progressive reinforcement fine-tuning method to further strengthen the alignment 
    through progressive high-level reward-based optimization.

    \item Experimental results demonstrate MindDriver achieves superior performance in both open-loop and closed-loop evaluations, as well as future frames generation, highlighting its effectiveness in reasoning and planning in autonomous driving. 

\end{itemize}
\section{Related Work}
\label{sec:related work}

\begin{figure*}[t]
    \centering
    \includegraphics[width=1.0\textwidth]{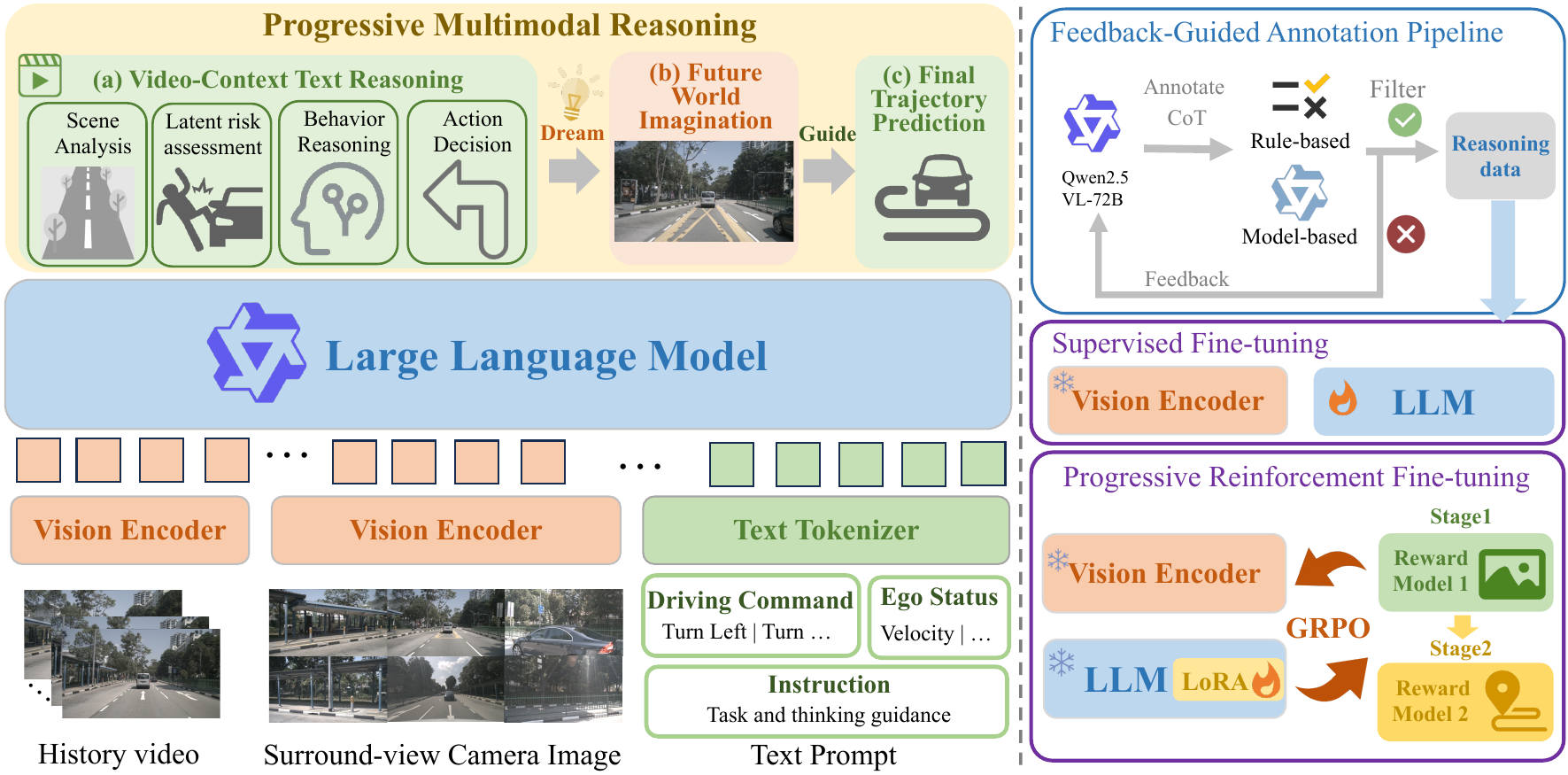}
    \caption{Overview. (Left) Framework of MindDriver. MindDriver conducts the perception-imagination-action process for accurate trajectory planning. (Right) (Top) Reasoning data annotation pipeline. The progressive multimodal reasoning data is auto-annotated by both rule-based and model-based filtering and feedback-guided regeneration. (Bottom) Progressive reinforcement fine-tuning is applied to enhance the progressive reasoning process.}
    \label{fig:overview}
\end{figure*}

\subsection{End-to-End Autonomous Driving}
End-to-end autonomous driving systems~\cite{jia2023think,  li2024hydra, sima2025centaur, liao2024diffusiondrive, li2024pretrain, zheng2024gaussianad, wang2025generativeaiautonomousdriving,zeng2024priordrive} directly map raw sensor inputs to driving trajectories through unified architectures, eliminating the need for hand-crafted intermediate modules. 
These approaches jointly optimize perception and planning, enabling task-coadapted feature abstraction and improved performance through gradient-based training.
Notable methods like UniAD \cite{hu2023planning} and VAD \cite{jiang2023vad} integrate perception, prediction, and planning into a single framework, improving open-loop benchmarks. SparseDrive \cite{sun2024sparsedrive} uses sparse perception to unify detection, tracking, and mapping, while GenAD \cite{zheng2024genad} and GoalFlow \cite{xing2025goalflowgoaldrivenflowmatching} employ generative models to predict multimodal trajectories.
However, these imitation learning-based systems struggle with interpretability and generalization in long-tail closed-loop scenarios \cite{tian2025thinktwiceenhancingllm,renz2025simlingovisiononlyclosedloopautonomous}. 
In this work, we develop a system excelling in both open and closed-loop evaluations while ensuring robust generalization.

\subsection{MLLM for Autonomous Driving}
MLLMs demonstrate strong contextual understanding and world knowledge~\cite{liu2025reasonplan, shao2023lmdriveclosedloopendtoenddriving, nie2024reason2driveinterpretablechainbasedreasoning,yuan2025unimapgen,liang2025persistent}, attracting growing integration with autonomous driving tasks. Methods like DriveVLM~\cite{DriveVLM} use a dual-system architecture where an LLM predicts trajectory primitives refined by an end-to-end model, while DriveLM~\cite{sima2024drivelm} employs visual question answering (VQA) for trajectory planning.
However, aligning semantic reasoning with precise action execution remains challenging. Solutions such as EMMA~\cite{hwang2024emma} incorporate hierarchical Chain-of-Thought for structured semantic reasoning, AutoVLA~\cite{zhou2025autovla} adopts adaptive reasoning for diverse scenarios, and FSDrive~\cite{zeng2025futuresightdrivethinkingvisuallyspatiotemporal} generates future scene images to enable visual imagination in trajectory planning.
To address these limitations, we propose a unified framework for progressive multimodal reasoning, enabling decision-oriented scene understanding and future image imagination to generate smooth, physically plausible trajectories through coherent multimodal inference.

\subsection{Reinforcement Fine-tuning}
DeepSeek-R1 \cite{guo2025deepseek} has confirmed that RFT \cite{ouyang2022training} exhibits significant potential in enhancing the performance and adaptability of VLMs. RAD \cite{gao2025rad} leverages 3D Gaussian splatting to conduct closed-loop Reinforcement Learning training. TrajHF \cite{li2025finetuning} adopts RFT technology to align trajectory generation models with safety constraints and human driving preferences. However, the application of RFT in end-to-end VLMs-based autonomous driving is still in its infancy~\cite{zheng2025driveagentr1advancingvlmbasedautonomous,yuan2025autodriver2incentivizingreasoningselfreflection}. Although some existing works use Gradient-Regularized Preference Optimization (GRPO) to help models improve trajectory planning capabilities, they can only reward the final outcome and cannot effectively optimize the intermediate process. In this work, we propose a progressive reinforcement fine-tuning approach that effectively rewards the process and achieves alignment in multimodal reasoning process. We apply RFT to the VLM framework, enhancing scene understanding capabilities and future image imagination abilities, while leveraging GRPO to ensure faster convergence and more stable training dynamics.

\begin{figure*}[t]
    \centering
    \includegraphics[width=1.0\textwidth]{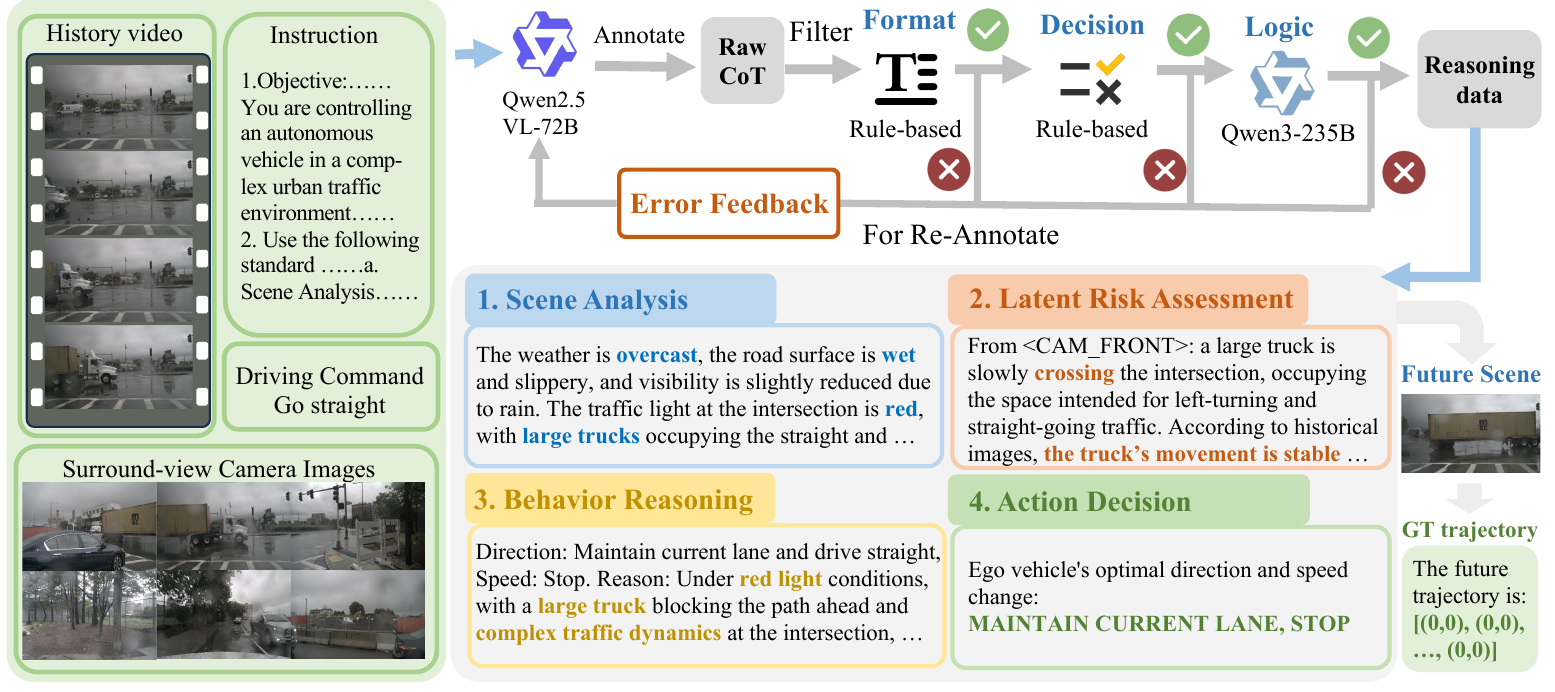}
    \caption{Auto-annotation pipeline for progressive multimodal reasoning training data. Qwen2.5-VL-72B first annotates raw CoT, which is then filtered based on format, decision, and logic. Failed cases are re-annotated using error feedback to improve generation quality. }
    \label{fig:cot}
\end{figure*}
\section{MindDriver}
\label{sec:method}

We propose MindDriver, a framework featuring a novel progressive multimodal reasoning method that enables smooth reasoning from semantic text understanding, future scene imagination, to physical trajectory prediction. 
To ensure reasoning alignment in MindDriver, we propose a feedback-guided auto-annotation pipeline and progressive reinforcement fine-tuning post-training strategy, which ensure strictly aligned training data and a consistent training process, respectively.

\subsection{Progressive Multimodal Reasoning Framework}
Following previous work~\cite{zeng2025futuresightdrive,liu2025reasonplan}, MindDriver
adopts camera images and text prompts as inputs, while conducting our proposed progressive multimodal reasoning, through a unified text reasoning and visual generation model.

\textbf{Model Inputs.} 
MindDriver ingests the temporal visual inputs, high-level driving commands, ego-vehicle states, and language instructions to perform driving reasoning and trajectory planning. 
The visual inputs comprise current images from six surround-view RGB cameras, along with four recent front-view frames as the history video, to capture scene dynamics without incurring substantial computational overhead. 
Additionally, following previous work~\cite{zeng2025futuresightdrive,wang2024omnidrive}, high-level driving commands (e.g, Turn Left, Turn Right), and ego vehicle status (i.e., current velocity and acceleration) are added. 
Finally, language instruction is utilized to structure these inputs into a prompt format that a large language model (LLM) can readily interpret.

\textbf{Progressive Multimodal Reasoning.} 
Most previous VLM-based methods focus on text reasoning~\cite{zhou2025autovla}, suffering from space misalignment between text semantic space and trajectory physical space. While recent work~\cite{zeng2025futuresightdrive} attempts to address this issue by using image reasoning instead of text as the intermediary, leveraging the semantic and physical information inherent in images, it lacks clear planning-oriented objective guidance, leaving the model uncertain about which objects to prioritize.
To address these, 
inspired by the human perception-imagination-action mechanism, 
we propose a progressive multimodal reasoning method that enables smooth reasoning from 
text understanding, through intermediate future image, to physical trajectories.
As shown in Fig.~\ref{fig:overview},
our method first leverages LLM's world knowledge to conduct semantic text reasoning, analyzing the scene, latent risk, and the action.
Then, guided by the planning-oriented analysis in text reasoning, our method imagines the future world scene, containing the analyzed moving tendency of each critical object and physical scene details.
Finally,
based on the dreamed image, our method predicts the final future trajectory with a coherent mapping from the physical details in image to physically-grounded trajectory outputs. 
  This end-to-end analysis ensures a progressive and smooth reasoning, enabling effective and interpretable autonomous driving planning.

\textbf{Unified Text Reasoning and Visual Generation.} 
To support our proposed progressive multimodal reasoning generation, inspired by recent works~\cite{Wang2024ILLUMEIY,wu2024liquid}, we unify textual reasoning and vision generation in a single LLM. 
Following~\cite{zeng2025futuresightdrive}, we expand the visual codebook of VQ-VAE~\cite{vqvae} to the vocabulary of the large language model, enabling MindDriver to generate discrete vision tokens. For visual generation, we utilize the tokenizer of VQ-VAE to encode images into discrete indices, and supervise the token prediction at each location for both modalities with a shared prediction head in LLMs.
Therefore, MindDriver adopts the general Language Modeling objective to directly maximize the likelihood of each multimodal sequence in an auto-regressive manner:
\begin{equation}\mathcal{L}=-\sum_{i=1}\log P_\theta(y_i|y_{< i}),\end{equation}
where $y_i$ denotes the text or visual token, and \( \theta \) is the LLM parameters. In inference, the VQ-VAE decoder (detokenizer) maps the vision tokens back to image pixels.

\subsection{Feedback-Guided Data Auto-annotation}

To produce reasonable and aligned progressive multimodal reasoning training data, we propose a video-context text reasoning format and a feedback-guided auto-annotation pipeline. 
The text reasoning data is merged with future scene image and trajectory to create complete reasoning data.
Finally, supervised fine-tuning (Fig.~\ref{fig:overview} right) is applied to the synthesized reasoning data to enable MindDriver with initial progressive multimodal reasoning ability.

\textbf{Video-Context Text Reasoning Format.}
We do not use existing public autonomous driving text reasoning data, as most of them are image-context~\cite{liu2025reasonplan,tian2024tokenize}, generating CoT from single-frame camera images.
Static inputs lack the motion tendencies of objects, which can lead to incorrect decisions, even for humans.
Although AutoVLA~\cite{zhou2025autovla} adopts a video-based CoT, it uses only the three front views for trajectory planning and ignores back views, increasing the risk from the back environment. 
To address these limitations, inspired by human driving logic, we design a video-context text reasoning, as shown in Fig.~\ref{fig:cot}. 
It consists of 
scene analysis, latent risk assessment, behavior reasoning, and action decision. 
The scene analysis and latent risk assessment are conducted based on existing camera images and the history video, which better captures the dynamics of each object in motion.
Action decision outputs the high-level driving decision, including direction and speed adjustments. Candidate decision categories are listed in the \textit{suppl.}

\textbf{Feedback-Guided Auto-annotation Pipeline.}
We develop a general pipeline for high-quality progressive multimodal reasoning data annotation.
As shown in Fig.~\ref{fig:cot}, this pipeline includes three filtering processes to control the data quality across different aspects, along with a feedback-guided re-annotation strategy to iteratively refine filtered samples.
Given inputs with current camera, history video, driving command, and instruction, the powerful MLLM Qwen2.5-VL-72B~\cite{Qwen2.5-VL} first generates a raw text CoT.
A rule-based format filter then checks structural completeness.
Next, a decision filter checks correctness by comparing generated action to GT decision derived from GT trajectory (details in \textit{suppl.}).
Finally, the logic filter evaluates reasoning soundness. Instead of reusing Qwen2.5-VL-72B, we employ the more advanced text-LLM Qwen3-235B-A22B-Instruct~\cite{yang2025qwen3} for robust logical validation and overcoming self-checking bias~\cite{yuan2025benchmarking}.
If any filter fails, error feedback is returned as context to improve re-annotation.
Feedback includes: (1) Format mismatches,
(2) Incorrect decisions vs. GT decisions, and (3) Logic errors summarized by Qwen3.
This feedback is combined with the raw CoT as input context for the next iteration.  
The pipeline details are listed in \textit{suppl.}
After that, the text CoT is concatenated with the ground truth future scene image and the trajectory with special tokens (<think>,<dream>,<answer>) to distinguish them, creating multimodal reasoning data.

\subsection{Progressive Reinforcement Fine-tuning}
Standard SFT has limitations in training our multimodal reasoning data for its token-level equally important supervision.
In multimodal reasoning, uniform token weighting can bias the model toward producing fluent text rather than maintaining a balanced understanding of both textual and visual information.
To enhance the reasoning ability of MindDriver, we introduce a progressive reinforcement fine-tuning post-training scheme, including two-stage learning with different rewards to optimize task-level behavior rather than token likelihoods.
This process progressively improves the dreamed image and the planned trajectory.

\textbf{Stage1: Dream Semantically Consistent Image.} This stage improves the model’s ability to generate a semantically consistent image based on the preceding text reasoning, compared to GT image.
Rather than optimizing pixel-level fidelity, we prioritize semantic consistency with the GT image, as the text CoT provides semantic guidance and highlights the approximate placement of key entities, which are crucial for downstream decisions.
To achieve this, we adopt the 
CLIP~\cite{clip} similarity to capture high-level semantic alignment between predicted and GT images, encouraging the model to dream images that preserve critical objects (e.g., traffic lights, pedestrians) in semantically correct locations.
The reward of stage 1 ($r_{Img}$) is formulated as:
\begin{equation}
    r_{Img} = \frac{E_{\text{CLIP}}({I_{\text{dream}}}) \cdot E_{\text{CLIP}}({I_{\text{GT}}})}{\lVert E_{\text{CLIP}}({I_\text{dream}}) \rVert  \lVert E_{\text{CLIP}}({I_\text{GT}}) \rVert}
\end{equation}
where $I_{\text{GT}}$ denotes the GT image, and $I_{\text{dream}}$ is the image generated by the MindDriver based on its preceding text CoT, $E_{\text{CLIP}}(\cdot)$ denotes the image encoder of CLIP.

\begin{table*}[t]
    \centering
    \caption{End-to-end trajectory planning experiments on nuScenes~\cite{Caesar2019nuScenesAM}. We evaluated the L2 and collision metrics based on the distinct computational methodologies of ST-P3~\cite{hu2022stp3} and UniAD~\cite{hu2023_uniad}, respectively. * indicates that ego status is additionally used. VAD~\cite{jiang2023vad} and UniAD~\cite{hu2023_uniad} results are derived from BEV-Planner~\cite{bev-planner}. $^{\dag}$ indicates the re-implemented results using official codes for Qwen2.5-VL-3B. 
    }
    \vspace{-1 em}
    \resizebox{\textwidth}{!}{ 
    \begin{tabular}{lcccccccc|cccccccc|c}
    \toprule
     \multirow{4}{*}{\textbf{Method}} & \multicolumn{8}{c}{\textbf{ST-P3 metrics}} & \multicolumn{8}{c}{\textbf{UniAD metrics}} & \multirow{4}{*}{{\textbf{LLM}}}  \\
    \cmidrule(lr){2-9} \cmidrule(lr){10-17}
    & \multicolumn{4}{c}{\textbf{L2 (m)} $\downarrow$} & \multicolumn{4}{c}{\textbf{Collision (\%)} $\downarrow$} & \multicolumn{4}{c}{\textbf{L2 (m)} $\downarrow$} & \multicolumn{4}{c}{\textbf{Collision (\%)} $\downarrow$} &   \\
    \cmidrule(lr){2-5} \cmidrule(lr){6-9} \cmidrule(lr){10-13} \cmidrule(lr){14-17}
    & 1s & 2s & 3s & \cellcolor{gray!20}Avg. & 1s & 2s & 3s & \cellcolor{gray!20}Avg. & 1s & 2s & 3s & \cellcolor{gray!20}Avg. & 1s & 2s & 3s & \cellcolor{gray!20}Avg. &  \\
    \midrule
    \multicolumn{18}{c}{\textbf{With Ego status methods}} \\
    \midrule
ST-P3*~\pub{ECCV22}~\cite{hu2022stp3} & 1.33 & 2.11 & 2.90 &   \cellcolor{gray!30}2.11 & 0.23 & 0.62 & 1.27 &   \cellcolor{gray!30}0.71 & - & - & - &   - & - & - & - &   - & -  \\ 
    VAD*~\pub{ICCV23}~\cite{jiang2023vad} & 0.17 & 0.34 & 0.60 &   \cellcolor{gray!30}0.37 & 0.04 & 0.27 & 0.67 &   \cellcolor{gray!30}0.33 & - & - & - &   - & - & - & - &   - & -  \\
    UniAD*~\pub{CVPR23}~\cite{hu2023_uniad} & - & - & - & - & - & - & - & - & \textbf{0.20} & \textbf{0.42} & \textbf{0.75} &   \cellcolor{gray!30}\textbf{0.46} & 0.02 & 0.25 & 0.84 &   \cellcolor{gray!30}0.37 & -  \\
    
    RDA-Driver*~\pub{ECCV24}~\cite{RDA-Driver} & 0.17 & 0.37 & 0.69 &   \cellcolor{gray!30}0.40 & 0.01 & \textbf{0.05} & 0.26 &   \cellcolor{gray!30}\textbf{0.10} & 0.23 & 0.73 & 1.54 &   \cellcolor{gray!30}0.80 & \textbf{0.00} & 0.13 & 0.83 &   \cellcolor{gray!30}0.32 & LLaVA-7B  \\

    BEV-Planner*~\pub{CVPR24}~\cite{bev-planner} & 0.16 & 0.32 & 0.57 &   \cellcolor{gray!30}0.35 & \textbf{0.00} & 0.29 & 0.73 &   \cellcolor{gray!30}0.34 & - & - & - &   - & - & - & - &   - & -  \\

    OminiDrive*~\pub{CVPR25}~\cite{Wang2024OmniDriveAH} & \textbf{0.14}  & 0.29 & 0.55 & \cellcolor{gray!30}0.33 & \textbf{0.00} & 0.13 & 0.78 & \cellcolor{gray!30}0.30  & - & - & - & - & - & - & - & - & LLaVA-7B \\

    FSDrive*$^{\dag}$~\pub{NeurIPS25}~\cite{zeng2025futuresightdrive} & 0.17 & 0.33 & 0.56 & \cellcolor{gray!30}0.35 & 0.07 & 0.10 & 0.24 & \cellcolor{gray!30}0.14 & 0.22 & 0.59 & 1.19 & \cellcolor{gray!30}0.67 & 0.07 & 0.14 & 0.75 & \cellcolor{gray!30}0.32  & Qwen2.5-VL-3B  \\ 
    
    Auto-VLA*~\pub{NeurIPS25}~\cite{zhou2025autovla} & 0.25  & 0.46 & 0.73 & \cellcolor{gray!30}0.48 & 0.07 & 0.07 & 0.26 & \cellcolor{gray!30}0.13 & 0.33 & 0.81 & 1.45 & \cellcolor{gray!30}0.86 & 0.08 & 0.11 & 0.85 & \cellcolor{gray!30}0.35 & Qwen2.5-VL-3B \\

    \midrule
    \textbf{MindDriver* (ours)} &  0.16 & \textbf{0.31} & \textbf{0.52} & \cellcolor{gray!30}\textbf{0.33} & 0.05 & 0.09 & \textbf{0.22} & \cellcolor{gray!30}0.12 & 0.22 & 0.57 & 1.15 & \cellcolor{gray!30}0.65 & 0.03 & \textbf{0.10} & \textbf{0.48} & \cellcolor{gray!30}\textbf{0.20}  & Qwen2.5-VL-3B  \\ 
    \midrule
    \multicolumn{18}{c}{\textbf{Without Ego status methods}} \\
    \midrule
        VAD~\pub{ICCV23}~\cite{jiang2023vad} & 0.69 & 1.22 & 1.83 &   \cellcolor{gray!30}1.25 & 0.06 & 0.68 & 2.52 &   \cellcolor{gray!30}1.09 & - & - & - &   - & - & - & - &   - & -  \\
     UniAD~\pub{CVPR23}~\cite{hu2023_uniad} & - & - & - & - & - & - & - & - & 0.59 & 1.01 & \textbf{1.48} &   \cellcolor{gray!30}1.03 & 0.16 & 0.51 & 1.64 &   \cellcolor{gray!30}0.77 & -  \\
     ELM~\pub{ECCV24}~\cite{zhou2024embodied}& - & - & - &   - & - & - & - & -&  \textbf{0.34}  & 1.23 & 2.57 & \cellcolor{gray!30}1.38 & 0.12 & 0.50 & 2.36 & \cellcolor{gray!30}0.99 & BLIP2-2.7B\\

    OccWorld~\pub{ECCV24}~\cite{zheng2023occworld} & 0.39  & 0.73 & 1.18 & \cellcolor{gray!30}0.77 & 0.11 & 0.19 & 0.67 & \cellcolor{gray!30}0.32 & 0.52  & 1.27 & 2.41 & \cellcolor{gray!30}1.40 & 0.12 & 0.40 & 2.08 & \cellcolor{gray!30}0.87 &GPT3-like\\

    BEV-Planner~\pub{CVPR24}~\cite{bev-planner} & 0.30 & 0.52 & 0.83 &   \cellcolor{gray!30}0.55 & 0.10 & 0.37 & 1.30 &   \cellcolor{gray!30}0.59 & - & - & - &   - & - & - & - &   - & -  \\
    OminiDrive~\pub{CVPR25}~\cite{Wang2024OmniDriveAH} & 0.40  & 0.80 & 1.32 & \cellcolor{gray!30}0.84 & \textbf{0.04} & 0.46 & 2.32 & \cellcolor{gray!30}0.94 & - & - &  - & - & - & - & - & - & LLaVA-7B \\

    PreWorld~\pub{ICLR25}~\cite{li2025semi}& - & - & - &   - & - & - & - & - & 0.49  & 1.22 & 2.32 & \cellcolor{gray!30}1.34 & 0.19 & 0.57 & 2.65 & \cellcolor{gray!30}1.14&- \\
    FSDrive$^{\dag}$~\pub{NeurIPS25}~\cite{zeng2025futuresightdrive} & 0.28 & 0.53 & 0.85 & \cellcolor{gray!30}0.55 & 0.07 & 0.15 & 0.40 & \cellcolor{gray!30}0.21 & 0.39 & 0.95 & 1.68 & \cellcolor{gray!30}1.01 & 0.05 & 0.25 & 1.03 & \cellcolor{gray!30}0.44  & Qwen2.5-VL-3B  \\ 
        \midrule
    \textbf{MindDriver (ours)}  & \textbf{0.27} & \textbf{0.51} & \textbf{0.82} & \cellcolor{gray!30}\textbf{0.53} & 0.06 & \textbf{0.12} & \textbf{0.32} & \cellcolor{gray!30}\textbf{0.17} & 0.35 & \textbf{0.87} & 1.57 & \cellcolor{gray!30}\textbf{0.93} & \textbf{0.04} & \textbf{0.19} & \textbf{0.91} & \cellcolor{gray!30}\textbf{0.38} & Qwen2.5-VL-3B  \\

    \bottomrule
    \end{tabular}}
    \label{tab:nusence}
\end{table*}

\textbf{Stage2: Predict Precise Trajectory.}
Building on the enhanced future-image imagination achieved in Stage 1, Stage 2 aligns the model with the trajectory-planning objective. 
Unlike SFT, which frames trajectory prediction as token prediction, this stage regulates trajectory using an L2 geometric distance, offering a more accurate supervised method for trajectory planning. 
Reward of stage 2 ($r_{\text{L2}}$) is:
\begin{equation}
    r_{\text{L2}} = \frac{\lambda - \mathrm{ADE}}{\alpha},\quad \mathrm{ADE} = \frac{1}{T}\sum_{t=1}^{T} \lVert \hat{y}_t - y_t \rVert_2
\end{equation}
where $\lambda$ denotes the maximum displacement error, and $\alpha$ is scaling factor to normalize the reward. The planning trajectory $\hat{y}_t$ is evaluated against the GT trajectory $y$, and ADE is computed as the average L2 distances over T time steps.

Similar to~\cite{zhou2025autovla}, we employ the GRPO algorithm \cite{shao2024deepseekmath}, which stabilizes training and improves convergence efficiency. 
Given a scenario input query $q$, comprising sensor images, the ego vehicle's state, and driving instruction, we sample a set of $G$ candidate outputs $O = \{o_1, o_2, \ldots, o_G\}$ from the old policy $\pi_{\theta_{\text{old}}}$. The current policy $\pi_{\theta}$ is then optimized using the normalized group-relative advantage $A_i$, by maximizing the following objective:
\begin{equation}
\mathcal{J}_{\text{GRPO}}(\theta) = \mathbb{E}_{q, o_i \sim \pi_{\theta_{\text{old}}}} \left[
\frac{1}{G} \sum_{i=1}^{G} \left( \mathcal{J}^{R}_i - \beta \, \mathbb{D}_{\text{KL}}(\pi_\theta \| \pi_{\text{ref}}) \right)
\right],
\end{equation}
\begin{equation}
\mathcal{J}_i^{R} = \min \left( \rho_i A_i,\ \text{clip} \left( \rho_i, 1 - \epsilon, 1 + \epsilon \right) A_i \right),
\end{equation}
\begin{equation}
\rho_i = \dfrac{\pi_\theta(o_i|q)}{\pi_{\theta_{\text{old}}}(o_i|q)}, \quad A_i = \frac{r_i - \text{mean}(\{r_j\}_{j=1}^G)}{\text{std}(\{r_j\}_{j=1}^G)}.
\end{equation}
\begin{equation}
r^{(s)} = 
\begin{cases}
r_{\text{Img}} + \lambda_1 \cdot r_{\text{format}}, & s = 1 , \\
r_{\text{L2}} + \lambda_2 \cdot r_{\text{format}}, & s = 2 .
\end{cases}
\end{equation}
where $\theta$ and $\theta_{old}$ denote the current and old policy parameters, $r_i$ is the reward for sample $o_i$, $r_{format}$ is the format reward, $\epsilon$ and $\beta$ are hyperparameters controlling the clipping range and the weight of the KL divergence regularization term, and $\pi_{\text{ref}}$ is the reference policy from the SFT stage.

\begin{table*}[!t]\small
\begin{center}
\caption{Future frames generation results on the nuScenes~\cite{Caesar2019nuScenesAM} dataset.
}
\vspace{-1 em}
\resizebox{\textwidth}{!}{ 
\begin{tabular}{l|ccccccc|c}
\toprule

\multirow{2}{*}{\textbf{Method}}&DriveGAN~\cite{kim2021drivegan}&DriveDreamer~\cite{wang2023drivedreamer}&Drive-WM~\cite{wang2023driving}&GenAD~\cite{yang2024genad} & GEM~\cite{Hassan2024GEMAG} &Doe-1~\cite{doe}& FSDrive~\cite{zeng2025futuresightdrive} & \multirow{2}{*}{\textbf{MindDriver (ours)}}   \\
&~\pub{CVPR21}&~\pub{ECCV24}&~\pub{CVPR24}&~\pub{CVPR24} & ~\pub{CVPR25}&~\pub{arxiv24}& ~\pub{NeruIPS25}&   \\
\midrule

\textbf{Type}&GAN&Diffusion &Diffusion&Diffusion&Diffusion&Autoregressive &Autoregressive &Autoregressive \\
\textbf{Resolution}  & 256$\times$256 & 128$\times$192  & 192$\times$384 & 256$\times$448 & 576$\times$1024& 384$\times$672&128$\times$192 & 128$\times$192\\

\midrule
\textbf{FID} $\downarrow$& 73.4& 52.6& 15.8& 15.4&10.5 &15.9& 10.1 & \textbf{9.4} \\

\bottomrule
\end{tabular} 
\label{tab:fid}}
\end{center}
\vspace{-1.5 em}

\end{table*}

\section{Experiments}
\label{sec:experiment}

\subsection{Experiment settings}
\textbf{Datasets.}
To comprehensively evaluate MindDriver's performance, we conducted both closed-loop and open-loop evaluations.
Following~\cite{jiang2023vad,zeng2025futuresightdrive}, we evaluate open-loop trajectory planning and future
frames generation on the nuScenes~\cite{caesar2020nuscenes}. The nuScenes contains 1,000 scenes of approximately 20 seconds each captured by six cameras providing 360-degree field of view.
Specifically, the dataset provides 28,130 (train) and 6,019 (val) samples.
Following~\cite{liu2025reasonplan,zhou2025autovla}, we evaluate the closed-loop driving performance on Bench2Drive~\cite{jia2024bench2drive},
which features challenging interactive scenarios based on the CARLA leaderboard v2. 
It provides an official training set where we use the base set (1000 clips)
for fair comparison with all the other baselines, which is divided into 950 clips for training and 50 clips for open-loop
validation. 
We evaluate the MindDriver on the official set of 220 short routes designed by Bench2Drive,

\textbf{Metrics.}
For open-loop evaluation, nuScenes includes L2 displacement error and collision rate for trajectory planning, following~\cite{hu2023_uniad,jiang2023vad}.
Notably, UniAD~\cite{hu2023_uniad} computes L2 metrics and collision rate at each timestep, whereas ST-P3~\cite{hu2022stp3} considers the average of all previous time-steps. 
We adopted both of these two different calculation methods. 
For future frames generation quality evaluation, we use Fréchet Inception Distance (FID)~\cite{fid}, following~\cite{zeng2025futuresightdrive,wang2023drivedreamer}.
For closed-loop evaluation, we adopt the metrics from~\cite{jia2024bench2drive}: 
(1) Driving Score (DS): overall performance metric; (2) Success Rate (SR): percentage of
infraction-free, timely completed episodes; (3) Efficiency (Effi): ego speed relative to neighboring
vehicles’ average; (4) Comfort (Comf): compliance with motion smoothness thresholds.

\textbf{Implementation.}
We employ Qwen2.5-VL-3B~\cite{Qwen2.5-VL} as our base VLM. During SFT, we use $1 \times 10^{-4}$ learning rate and batch size of 32, for 12 epochs (nuSences) and 6 epochs (Bench2Drive).
We extend the visual codebook of MoVQGAN~\cite{Zheng2022MoVQgan} to LLM vocabulary and use its detokenizer to map LLM-predicted visual tokens to pixel space.
During progressive RFT, we use $3 \times 10^{-6}$ learning rate and batch size 16, for 700 (stage 1) and 500 (stage 2) steps in nuSences, 1400 (stage 1) and 1000 (stage 2) steps in Bench2Drive. 
All the experiments were run on 16 Nvidia H20. Additional detailed information is listed in the \textit{suppl.}

\begin{table}[t]
  \caption{Closed-loop Results on the Bench2Drive (CARLA) Benchmark. *: the ego status is additionally use. $\ddagger$ : trained not on Bench2Drive training set}
  \vspace{-1 em}
  \label{tab:b2d}
  \centering
  \scriptsize
  \setlength{\tabcolsep}{6.5pt}
  \renewcommand{\arraystretch}{1.2}
  \begin{tabular}{l|cccc}
    \toprule
    \textbf{Method}                                      & \textbf{DS} $\uparrow$    &  \textbf{SR} (\%) $\uparrow$  &  \textbf{Effi} $\uparrow$   & \textbf{Comf} $\uparrow$ \\
    \midrule
    AD-MLP*~\pub{Arxiv23} \cite{zhai2023rethinking}            & 18.05           & 0.00                 &  48.45        & 22.63   \\
    UniAD-Base*~\pub{CVPR23} \cite{hu2023planning}            & 45.81           & 16.36                & 129.21        & 43.58   \\
    VAD*~\pub{ICCV25} \cite{jiang2023vad}                     & 42.35           & 15.00                & 157.94        & \textbf{46.01}    \\
    TCP-traj*~\pub{NeurIPS22} \cite{wu2022trajectory}            & 59.90           & 30.00                & 76.54         & 18.08 \\
    DriveAdapter*~\pub{CVPR23} \cite{jia2023driveadapter}     & 64.22           & 33.08                & 70.22         & 16.01  \\
    ReasonPlan*~\pub{CoRL25} \cite{liu2025reasonplan} &  64.01 & 34.55 & \textbf{180.64} & 25.63 \\
    \midrule
     AutoVLA*$\ddagger$~\pub{NeurIPS25} \cite{zhou2025autovla}                                   & \textcolor{gray}{78.84}  & \textcolor{gray}{57.73}      & \textcolor{gray}{146.93}        & \textcolor{gray}{39.33} \\
     \midrule
     \textbf{MindDriver(ours)}                                    & \textbf{65.48}  & \textbf{39.55}      & 143.21        & 34.63 \\
    \bottomrule
  \end{tabular}
\end{table}

\begin{table}[!t]\footnotesize
\begin{center}
\setlength{\tabcolsep}{3pt}
\renewcommand{\arraystretch}{1.2}
\caption{Ablation results of different CoT.
}
\vspace{-1 em}
\begin{tabular}{c|cccc|cccc}
\toprule

 \multirow{2}{*}{\textbf{Type}} &\multicolumn{4}{c|}{\textbf{L2} (m) $\downarrow$} & 
\multicolumn{4}{c}{\textbf{Collision} (\%) $\downarrow$} \\
& \textbf{1s} & \textbf{2s} & \textbf{3s} & \cellcolor{gray!30}\textbf{Avg.} & \textbf{1s} & \textbf{2s} & \textbf{3s} & \cellcolor{gray!30}\textbf{Avg.} \\

\midrule

None & 0.38 & 0.93 & 1.67 & \cellcolor{gray!30}0.99 & 0.13  & 0.35 & 1.18 & \cellcolor{gray!30}0.56 \\

Text CoT & 0.37 & 0.90 & 1.61 & \cellcolor{gray!30}0.96 & 0.08 & 0.23 & 1.05 & \cellcolor{gray!30}0.46\\

Image CoT & 0.39 & 0.98 & 1.81 & \cellcolor{gray!30}1.06 & 0.15 & 0.37 & 1.13 & \cellcolor{gray!30}0.55 \\

MultiModal(I2T) CoT & 0.38 & 0.95 & 1.70 & \cellcolor{gray!30}1.01 & 0.07 & 0.25 & 1.08 & \cellcolor{gray!30}0.47 \\

MultiModal(T2I) CoT & \textbf{0.36} & \textbf{0.89} & \textbf{1.60} & \cellcolor{gray!30}\textbf{0.95} & \textbf{0.05} & \textbf{0.22} & \textbf{0.97} & \cellcolor{gray!30}\textbf{0.41} \\

\bottomrule
\end{tabular} 
\label{tab:cot ablation}
\end{center}
\vspace{-2 em}
\end{table}

\subsection{Main results}

\textbf{Open-loop evaluation on nuSences.}
Tab.~\ref{tab:nusence} illustrates open-loop trajectory planning performance on nuScenes.
As for results without ego status, MindDriver outperforms previous SOTA methods on both ST-P3 and UniAD metrics, including non-autoregressive (e.g., UniAD) and autoregressive methods (e.g., OccWorld).
Notably, MindDriver's multimodal reasoning surpasses image-only CoT methods (e.g., FSDrive), indicating that incorporating textual reasoning before future image generation improves trajectory quality and reduces collisions.
Additionally, MindDriver exceeds text-only CoT approaches (e.g., AutoVLA), suggesting that subsequent future world dreaming is particularly effective at lowering collision rates.

\textbf{Evaluation of generated image.}
As shown in Tab.~\ref{tab:fid}, following prior work~\cite{zeng2025futuresightdrive,wang2023driving}, we report the FID of the generated image to validate their visual quality. To balance the quality and generation speed, we generate frames at 128x192 resolution. It is observed that MindDriver achieves superior FID, outperforming even specialized diffusion-based models (e.g., DriveDreamer, Drive-WM).
Compared with the image-only CoT method, FSDrive, MindDriver's lower FID indicates that prefixed textual reasoning enhances scene understanding and action decisions, leading to more accurate future image generation.

\textbf{Closed-loop evaluation on Bench2Drive (CARLA).}
For results in Tab.~\ref{tab:b2d}, AutoVLA is trained on a much larger set of datasets, including nuPlan~\cite{park2025nuplanqa}, CARLA-Garage~\cite{Jaeger_2023_ICCV}, and so on.
It is observed that MindDriver achieves competitive closed-loop performance compared to SOTA methods DriveAdapter~\cite{jia2023driveadapter}, which utilizes privileged expert feature distillation, and ReasonPlan~\cite{liu2025reasonplan}, which predicts future image before text reasoning.
MindDriver achieves a higher driving score and success rate (39.55\%), even without using ego-status.
This illustrates MindDriver's strong ability to reason across diverse driving scenes and validate its robustness under complex,
multi-intent scenarios.

\subsection{Ablation Study}

\begin{table}[!t]\footnotesize
\begin{center}
\setlength{\tabcolsep}{3pt}
\renewcommand{\arraystretch}{1.2}
\caption{Ablation results of dreaming different future scene.
}
\vspace{-1 em}
\begin{tabular}{c|cccc|cccc}
\toprule

 \multirow{2}{*}{\textbf{Type}} &\multicolumn{4}{c|}{\textbf{L2} (m) $\downarrow$} & 
\multicolumn{4}{c}{\textbf{Collision} (\%) $\downarrow$} \\
& \textbf{1s} & \textbf{2s} & \textbf{3s} & \cellcolor{gray!30}\textbf{Avg.} & \textbf{1s} & \textbf{2s} & \textbf{3s} & \cellcolor{gray!30}\textbf{Avg.} \\

\midrule

Next 1.5s image & 0.39  & 0.95 & 1.68 & \cellcolor{gray!30}1.01 & 0.08 & 0.30 & \textbf{0.97} & \cellcolor{gray!30}0.45 \\
Next 1s image & 0.40  & 0.96 & 1.75 & \cellcolor{gray!30}1.04 & \textbf{0.03} & 0.29 & \textbf{0.97} & \cellcolor{gray!30}0.43 \\
Next 0.5s image & \textbf{0.36} & \textbf{0.89} & \textbf{1.60} & \cellcolor{gray!30}\textbf{0.95} & 0.05 & \textbf{0.22} & \textbf{0.97} & \cellcolor{gray!30}\textbf{0.41} \\
\bottomrule
\end{tabular} 
\label{tab:frame ablation}
\end{center}
\vspace{-1 em}
\end{table}

\begin{figure*}[t]
    \centering
    \includegraphics[width=1.0\textwidth]{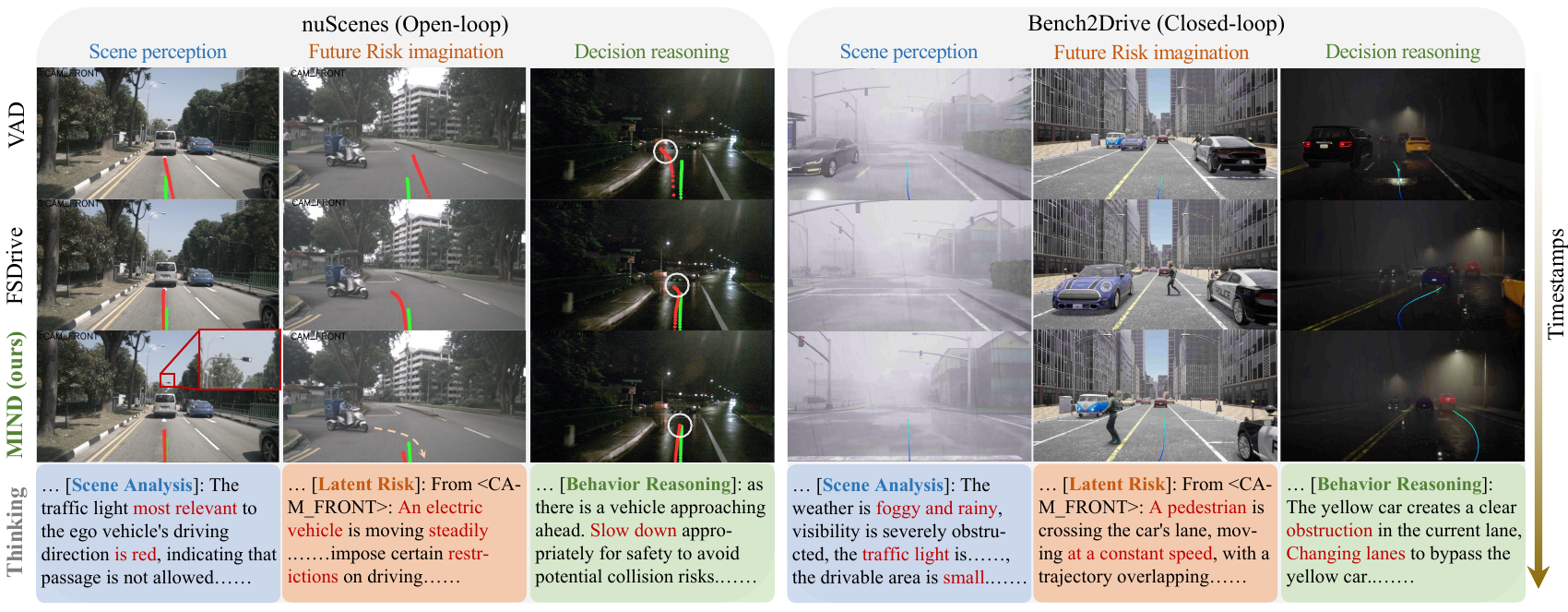}
    \vspace{-2 em}
    \caption{Qualitative comparison of MindDriver with baselines. (Left) Three scenarios from the open-loop nuScenes benchmark. The red trajectory is the prediction and the green one is the GT. (Right) The performance variation with timestamps on closed-loop Bench2Drive.} 
    \label{fig:example}
    \vspace{-1 em}
\end{figure*}

\begin{table}[!t]\footnotesize
\begin{center}
\setlength{\tabcolsep}{3pt}
\renewcommand{\arraystretch}{1.2}
\caption{Ablation results of data filtering.
}
\vspace{-1 em}
\begin{tabular}{c|cccc|cccc}
\toprule
 \multirow{2}{*}{\textbf{Strategy}} &\multicolumn{4}{c|}{\textbf{L2} (m) $\downarrow$} & 
\multicolumn{4}{c}{\textbf{Collision} (\%) $\downarrow$} \\
& \textbf{1s} & \textbf{2s} & \textbf{3s} & \cellcolor{gray!30}\textbf{Avg.} & \textbf{1s} & \textbf{2s} & \textbf{3s} & \cellcolor{gray!30}\textbf{Avg.} \\
\midrule

No CoT & 0.38 & 0.93 & 1.67 & \cellcolor{gray!30}0.99 & 0.13  & 0.35 & 1.18 & \cellcolor{gray!30}0.56 \\
\midrule

Raw CoT & 1.32 & 2.50 & 3.62 & \cellcolor{gray!30}2.48 & 0.17 & 1.06 & 2.97 & \cellcolor{gray!30}1.40 \\
+ 3 Filters & \textbf{0.37}  & 0.92 & 1.64 & \cellcolor{gray!30}0.98 & \textbf{0.08} & 0.32 & 1.18 & \cellcolor{gray!30}0.53 \\
+ Error feedback & \textbf{0.37} & \textbf{0.90} & \textbf{1.61} & \cellcolor{gray!30}\textbf{0.96} & \textbf{0.08} & \textbf{0.23} & \textbf{1.05} & \cellcolor{gray!30}\textbf{0.46} \\
\bottomrule
\end{tabular} 
\label{tab:data filter ablation}
\end{center}
\end{table}

\textbf{Different CoT Type.}
Tab.~\ref{tab:cot ablation} shows the ablation study on different CoT designs.
MultiModal(I2T/T2I) CoT denotes different text and image order in multimodal reasoning: I2T dreams an image first and then performs text reasoning, whereas T2I first conducts text reasoning and then dreams an image.
It is observed that pure text CoT provides clear improvement over the baseline, and our proposed multimodal CoT further enhances this by dreaming future scene images. Pure image CoT offers limited benefits due to the absence of planning-oriented reasoning.
Within multimodal reasoning, T2I consistently outperforms I2T (as in ReasonPlan [28]), which aligns with the logic of human reasoning: high-level semantic planning precedes accurate scene estimation for precise control.

\textbf{Dream Different Future Scene.}
We conduct ablation studies on different future frames generation in Tab.~\ref{tab:frame ablation}, covering the imagination of the next 0.5/1/1.5s scene image. 
Tab.~\ref{tab:frame ablation} shows that dreaming the next 0.5s scene image achieves the best performance.
This likely stems from better temporal alignment with the frame sampling interval (0.5s/frame) of the input history video, which simplifies prediction.
 Additionally, predictions over 1/1.5s span longer distances, increasing uncertainty and reducing accuracy, especially for rare or emergent events.

\textbf{Reasoning Data Filtering.}
Tab.~\ref{tab:data filter ablation} ablates our data-filtering strategy for text reasoning. 
Training with raw CoT annotated by Qwen2.5-VL-72B yields a significant decrease in performance than the No CoT baseline.
This indicates that the unfiltered CoT is low-quality and contains substantial logical and decision-making errors.
However, applying our three filters markedly improves CoT quality and creates an obvious performance improvement over the baseline. 
Additionally, our proposed error feedback-guided strategy further improves the CoT quality by prompting the VLM with history error feedback, producing more accurate reasoning, and achieving the best performance.

\begin{table}[!t]\footnotesize
\begin{center}
\setlength{\tabcolsep}{2.7pt}
\renewcommand{\arraystretch}{1.2}
\caption{Ablation results of Progressive RFT.
}
\vspace{-1 em}
\begin{tabular}{c|cccc|cccc|c}
\toprule
 \multirow{2}{*}{\textbf{Type}} &\multicolumn{4}{c|}{\textbf{L2} (m) $\downarrow$} & 
\multicolumn{4}{c|}{\textbf{Collision} (\%) $\downarrow$} & \textbf{FID} $\downarrow$\\
& \textbf{1s} & \textbf{2s} & \textbf{3s} & \cellcolor{gray!30}\textbf{Avg.} & \textbf{1s} & \textbf{2s} & \textbf{3s} & \cellcolor{gray!30}\textbf{Avg.} & \cellcolor{gray!30}Score \\
\midrule
None RL & 0.36 & 0.89 & 1.60 & \cellcolor{gray!30}0.95 & 0.05 & 0.22 & 0.97 & \cellcolor{gray!30}0.41 & \cellcolor{gray!30}9.8 \\
One-stage RL & 0.37 & 0.91 & 1.64 & \cellcolor{gray!30}0.97 & 0.07 & 0.23 & 0.95 & \cellcolor{gray!30}0.42 & \cellcolor{gray!30}9.7 \\
Progressive RL & \textbf{0.35} & \textbf{0.87} & \textbf{1.57} & \cellcolor{gray!30}\textbf{0.93} & \textbf{0.04} & \textbf{0.19} & \textbf{0.91} & \cellcolor{gray!30}\textbf{0.38} & \cellcolor{gray!30}\textbf{9.4}\\
\bottomrule
\end{tabular} 
\label{tab:rl ablation}
\end{center}
\vspace{-2 em}
\end{table}

\textbf{Progressive RFT}.
Tab.\ref{tab:rl ablation} illustrates the ablation on RFT strategy.
It is observed that two-stage progressive RFT achieves the best performance, outperforming the one-stage variant that optimizes a weighted sum of image and trajectory rewards (with weights 0.33 and 0.67).
One-stage RFT shows a marginal difference with baseline, 
likely because jointly balancing image generation and trajectory prediction is difficult.
Meanwhile, our proposed progressive RFT first enhances the model to dream an accurate scene image, improving alignment between the text CoT and the image. The second stage further optimizes trajectory planning, 
further aligning generated images with the predicted trajectories.

\subsection{Qualitative Visualization}
Fig.~\ref{fig:example} presents the qualitative results of MindDriver in representative open-loop and closed-loop evaluation scenarios, illustrating its progressive reasoning process and the corresponding predicted trajectories. Compared with baseline methods, MindDriver demonstrates superior performance in scenarios involving complex environmental interactions and latent risk reasoning. In the open-loop nuScenes benchmark, with regard to perceptual capabilities, the MindDriver can detect the most relevant traffic lights, outperforms FSDrive significantly in scenarios involving dynamic obstacles. In the closed-loop Bench2Drive benchmark, MindDriver can accurately recognize relevant traffic lights even under low visibility conditions.

\section{Conclusion}
\label{sec:conclusion}

This paper proposes MindDriver, a unified framework based on progressive multimodal reasoning that enables VLMs to 
imitate human-like progressive thinking.
The framework first conducts semantic-space text reasoning to achieve comprehensive scene understanding.
Then, guided by this, it dreams the future scene image, and ultimately predicts the physical-space trajectory.
To ensure aligned multimodal reasoning, we introduce a feedback-guided data annotation pipeline. Furthermore, we develop a progressive reinforcement fine-tuning method to optimize the alignment through progressive high-level
reward-based learning. 
 Extensive experiments on both open-loop and closed-loop validate the effectiveness of MindDriver, advancing autonomous driving toward more reliable reasoning.

\textbf{Limitations and future work.} 
Although MindDriver achieves comparable inference speed (1 Hz) on an NVIDIA RTX 4090 GPU with other VLM-based methods, it is highly GPU-dependent, requiring significant computing.
Moreover, the current work only considers the generation of front-view images; future efforts could explore richer and more detailed visual outputs.

\clearpage
\setcounter{page}{1}
\maketitlesupplementary

\section{Reasoning Annotation Pipeline}
\begin{figure}[t]
    \centering
    \includegraphics[width=0.45\textwidth]{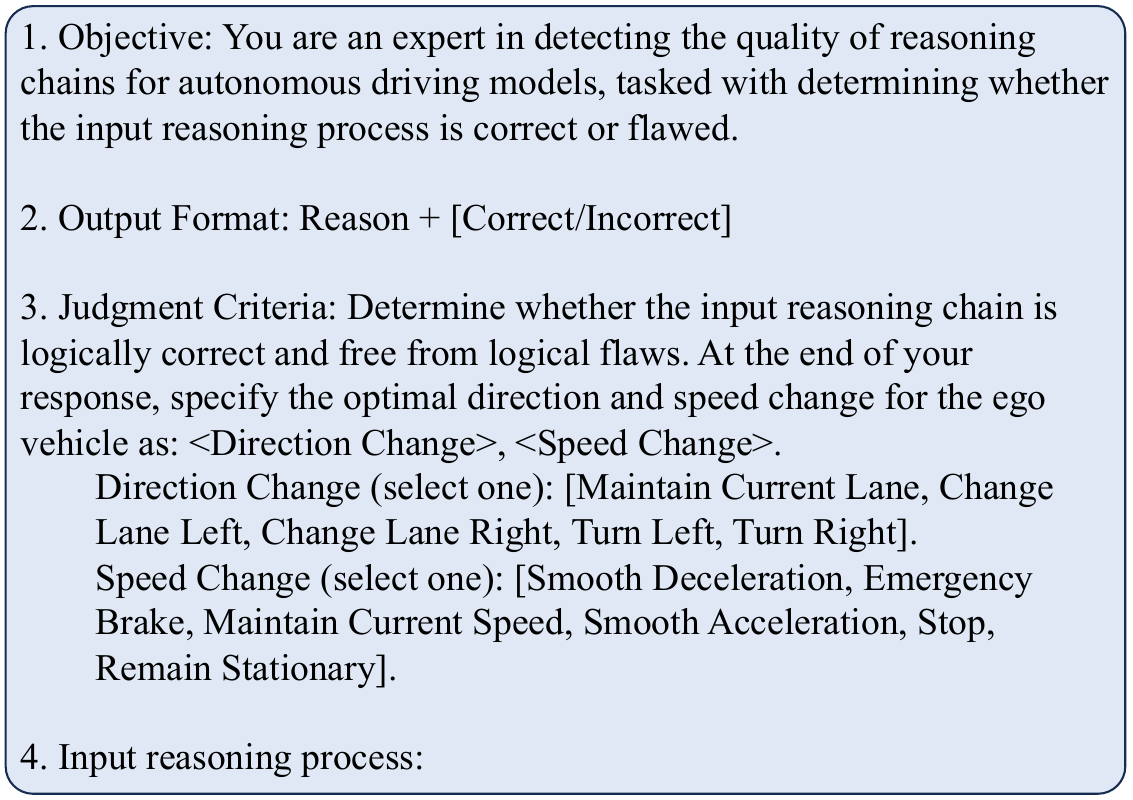} 
    \caption{Prompt for logical verification to Qwen3-235B-A22B-Instruct.}
    \label{fig:logic prompt} 
\end{figure}

We develop a general pipeline for high-quality progressive multimodal reasoning data annotation.
This pipeline includes three filtering processes to control the data quality across different aspects, along with a feedback-guided re-annotation strategy to iteratively refine filtered samples.
Given inputs with the current camera, history video, driving command, and instruction, the powerful MLLM Qwen2.5-VL-72B~\cite{Qwen2.5-VL} first generates a raw text CoT. The prompt for Qwen2.5-VL-72B is shown in Fig.~\ref{fig:qwen72b}. The text CoT then passes through three filters:
\begin{itemize}
\item \textbf{Format Filter:} This rule-based filter checks whether the text reasoning is composed of the four parts: (1) Scene Analysis, (2) Latent Risk Assessment, (3) Behavior Reasoning, and (4) Action Decision (including both direction and speed prediction).  
\item \textbf{Decision Filter:} It checks decision correctness by comparing the generated action to the GT decision derived from the GT trajectory.
The process for generating ground truth (GT) labels is as follows:
Leveraging statistical insights into dynamic vehicle parameters (e.g., velocity and acceleration) from the dataset, and informed by prior knowledge of real-world driving behaviors, we conducted clustering analysis on future vehicle trajectories. We experimented with different numbers of clusters (7, 10, and 49) to evaluate the effectiveness of trajectory pattern segmentation. The results revealed that smaller cluster counts led to highly imbalanced trajectory distributions, failing to capture the diversity of driving behaviors.
To enhance model learning and generalization, we adopted a fine-grained trajectory categorization strategy. Specifically, for accelerating and decelerating vehicles, we used the 30th and 60th percentiles of their acceleration distributions as thresholds to differentiate behavior subcategories. A similar percentile-based thresholding approach was applied to left-turning and right-turning vehicles, based on their turning dynamics. This method enables more discriminative and behaviorally meaningful trajectory labeling, thereby supporting more accurate prediction modeling. The final selected meta actions are shown in the Tab.~\ref{tab:meta}.

\begin{table}[t]\footnotesize
\centering
\setlength{\tabcolsep}{6pt} 
\renewcommand{\arraystretch}{1.2} 
\caption{Meta action type in desion filter.}
\begin{tabular}{c|m{0.3\textwidth}} 
\toprule
\textbf{Behavior} & \textbf{Meta Action Type} \\
\midrule
Direction Change & [Maintain Current Lane, Change Lane Left, Change Lane Right, Turn Left, Turn Right] \\
\midrule
Speed Change & [Smooth Deceleration, Emergency Brake, Maintain Current Speed, Smooth Acceleration, Stop, Remain Stationary] \\
\bottomrule
\end{tabular}
\label{tab:meta}
\end{table}
\item \textbf{Logic Filter:} This filter evaluates the reasoning soundness of CoT. Instead of reusing Qwen2.5-VL-72B, we employ the more advanced text-LLM Qwen3-235B-A22B-Instruct~\cite{yang2025qwen3} for robust logical validation and overcoming self-checking bias~\cite{yuan2025benchmarking}. The prompt of Qwen3-235B-A22B-Instruct is illustrated in Fig.~\ref{fig:logic prompt}. To enable better understanding, we show an example of this logical quality check in Fig.~\ref{fig:logic example}. Based on Qwen3's robust logical analysis capabilities, a critical examination of the preliminary response from Qwen2.5VL-72B identified a broken causal chain in its reasoning process. The reasoning incorrectly conflates two distinct operational phases, firstly, valid recommendation for post-green-light behavior ("safe passage after the light turns green"), which is contextually appropriate; secondly, mandatory red-light behavior (complete stop and wait), which was not explicitly specified. This conflation erroneously applies the speed-adjustment guidance for post-green-light conditions to current red-light state, resulting in a conclusion that is fundamentally disconnected from the actual traffic scenario. Therefore, by applying similar logical validation,  reasoning errors can be identified and corrected.

\end{itemize}

\begin{figure}[t]
    \centering
    \includegraphics[width=0.45\textwidth]{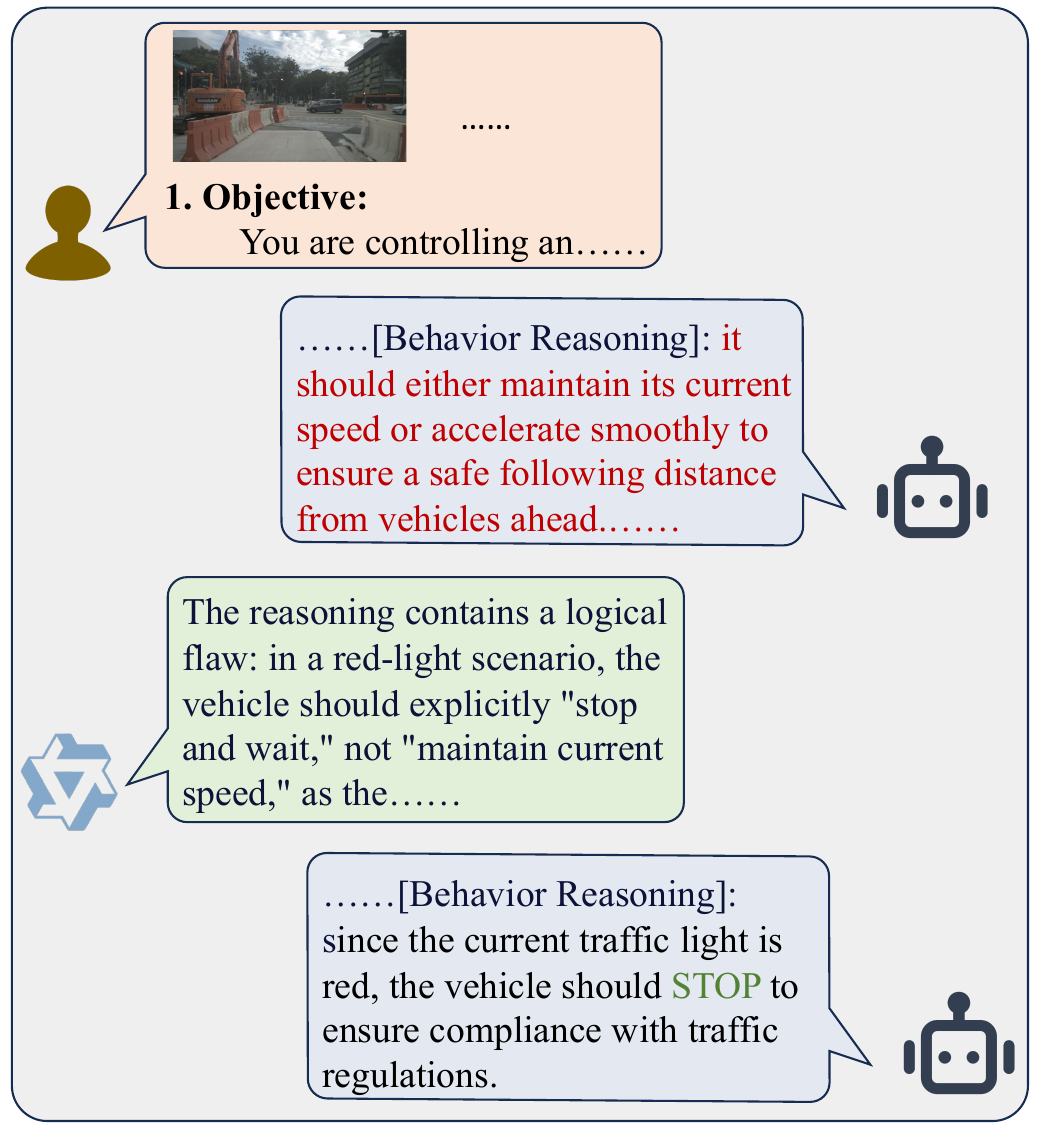} 
    \caption{Combined COT with feedback for re-annotation.}
    \label{fig:feedback example} 
\end{figure}

\textbf{Feedback-guided Re-annotation.}
If any above filter fails, error feedback is returned as context to improve re-annotation.
As shown in Tab.~\ref{tab:feedback}, the error feedback includes: (1) Format error: the detailed missing parts considering scene analysis, latent risk assessment, behavior reasoning, and action decision.
(2) Decision error: Incorrect decisions vs. GT decisions, for both direction and speed decisions, and (3) Logic error: return the summarized logic errors generated by Qwen3-235B-A22B-Instruct.
This feedback is combined with the raw CoT as input context for the next iteration. This process is shown in Fig.~\ref{fig:feedback example}.

\begin{figure*}[t]
    \centering
    \includegraphics[width=0.95\textwidth]{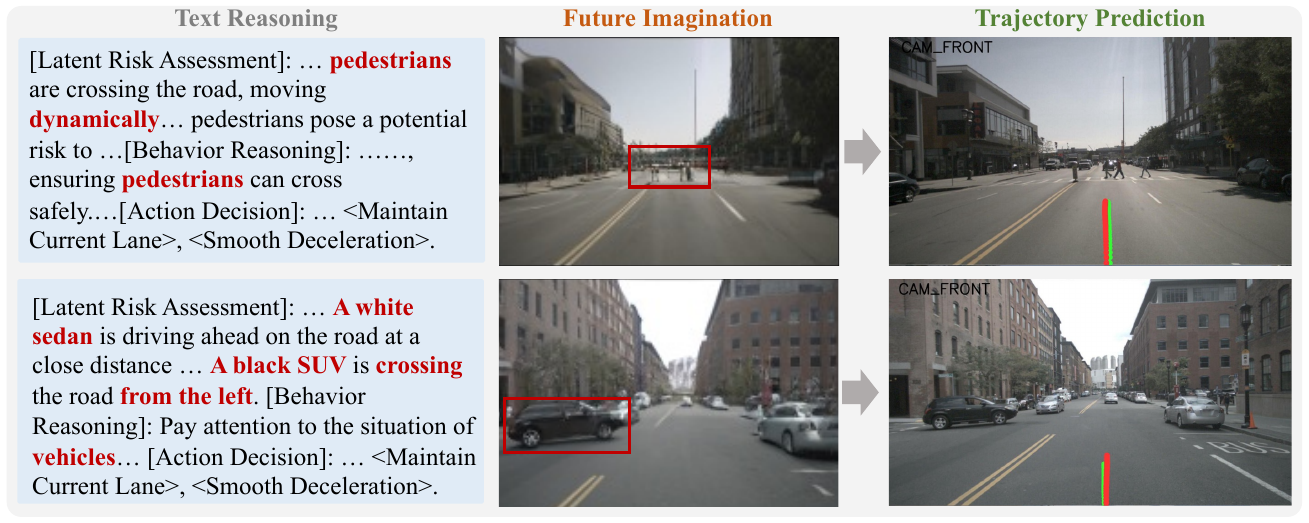}
    \caption{Qualitative Analysis of MindDriver. Red represents our predicted values, and green represents the ground truth (GT).}
    \label{figdingxing}
\end{figure*}

After that, the text CoT is concatenated with the ground truth future scene image and the trajectory with special tokens (<think>,<dream>,<answer>) to distinguish them, creating multimodal reasoning data. It is formatted as:

\begin{equation}
    \text{\textless t\textgreater\,Tok\textsubscript{Text CoT}\textless/t\textgreater
      \textless d\textgreater\,Tok\textsubscript{Img}\textless/d\textgreater
      \textless a\textgreater\,Tok\textsubscript{Traj}\textless/a\textgreater}
\end{equation}
where <t>, <d>, <a> denotes <think>,<dream>,<answer> special tokens. Tok\textsubscript{Text CoT}, Tok\textsubscript{Img}, and Tok\textsubscript{Traj} are the tokens of the text CoT, the dreamed image, and the predicted trajectory.

\begin{table}[t]\footnotesize
\centering
\setlength{\tabcolsep}{6pt} 
\renewcommand{\arraystretch}{1.2} 
\caption{Specific feedback type in data auto-annotation.}
\begin{tabular}{c|m{0.3\textwidth}} 
\toprule
\textbf{Error Type} & \textbf{Feedback Content Example} \\
\midrule
Format Error & Missing Scene Analysis / ...
Missing Action Decision part. \\
\midrule
Decision Error & 1.Direction decision error(GT: Change Lane Right; Prediction: Turn Right). 

2.Speed decision error(GT: Smooth Deceleration; Prediction: Maintain Current Speed). \\
\midrule
Logic Error & Reasoning is discontinuous ... consideration is incomplete \\
\bottomrule
\end{tabular}
\label{tab:feedback}
\end{table}

\begin{figure*}[t]
    \centering
    \includegraphics[width=0.95\textwidth]{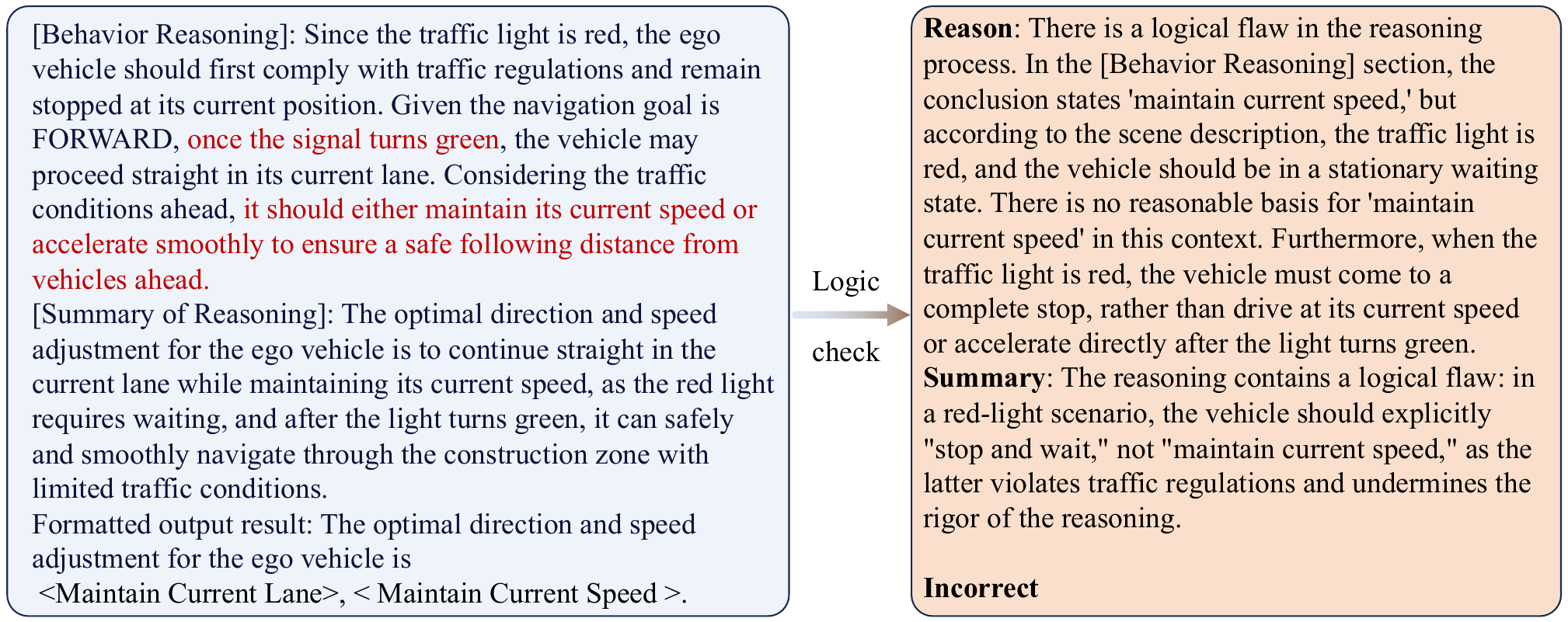} 
    \caption{An example of a logical quality check.}
    \label{fig:logic example} 
\end{figure*}

\begin{figure*}[t]
    \centering
    \includegraphics[width=0.95\textwidth]{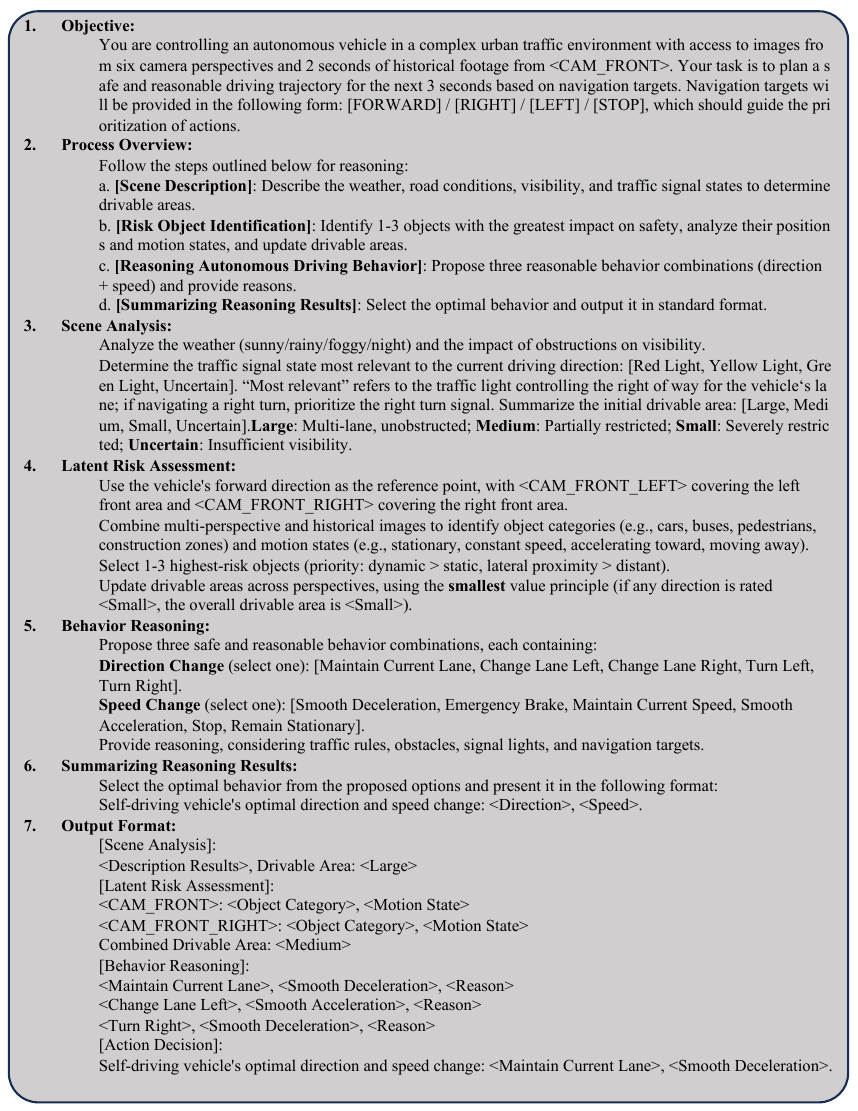} 
    \caption{Prompt for CoT annotation by Qwen2.5-VL-72B}
    \label{fig:qwen72b} 
\end{figure*}

\begin{figure*}[t]
    \centering
    \includegraphics[width=0.95\textwidth]{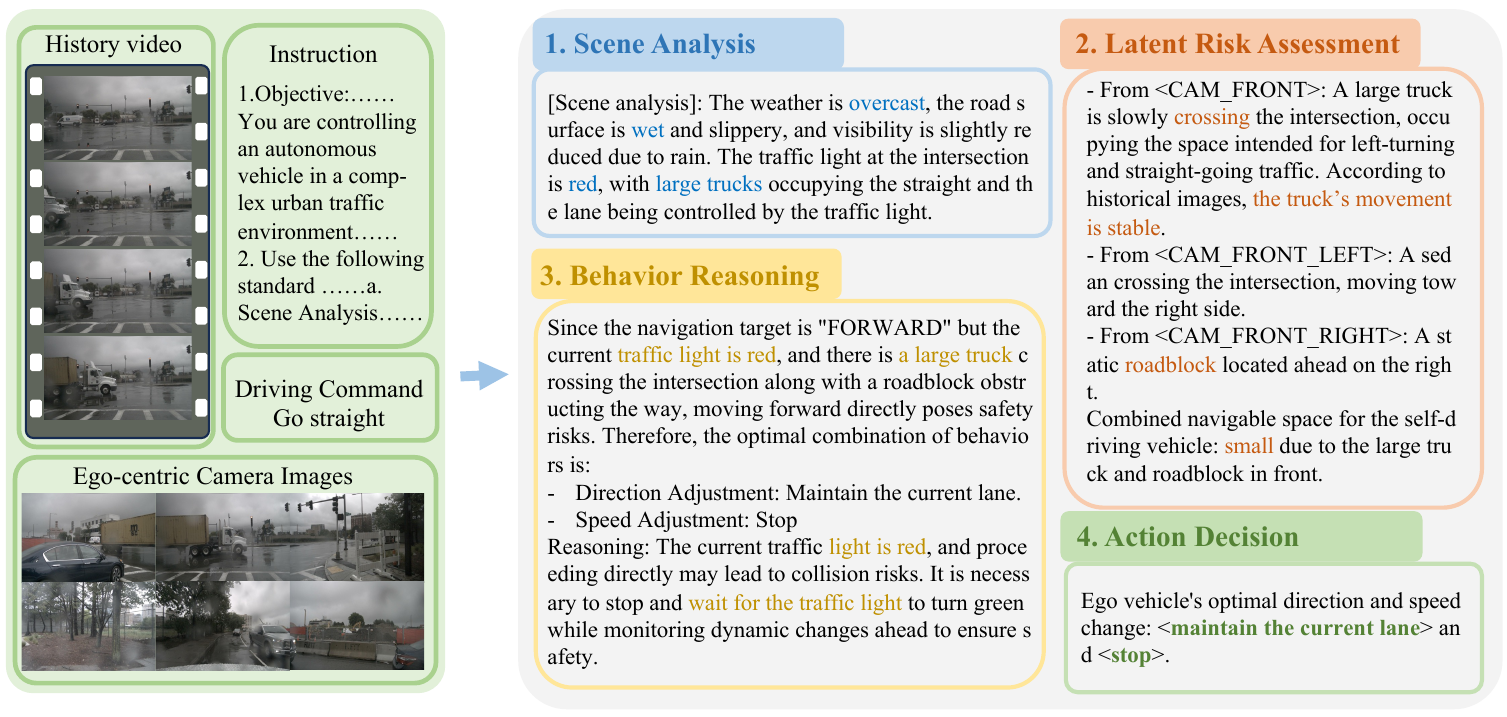} 
    \caption{A complete sample of the annotation dataset.}
    \label{fig:cot} 
\end{figure*}

\section{Implementation Details}
All experiments are conducted on 16 NVIDIA H20 GPUs (96 GB each). We employ Qwen2.5-VL-3B~\cite{Qwen2.5-VL} as our base VLM. During SFT, we use $1 \times 10^{-4}$ learning rate and batch size of 32, for 12 epochs (nuSences) and 6 epochs (Bench2Drive). 
For our feedback annotation pipline, We set the maximum number of iterative rounds to 3. The vision
encoder of MindDriver is frozen, and the LLM is fully fine-tuned in the SFT stage.
During progressive RFT, we use $3 \times 10^{-6}$ learning rate and batch size 16, for 700 (stage 1) and 500 (stage 2) steps in nuSences, 1400 (stage 1) and 1000 (stage 2) steps in Bench2Drive. 
We set $\lambda$ = 10 and $\alpha$ = 6 for L2 reward.
In Eq. 7 of main paper, $\lambda_1$ and $\lambda_1$ are both set as 10 for strict format learning in RFT. The format reward $r_{\text{format}}$ is to check whether the answer includes 6 parsed points (If yes, set 1; Otherwise set 0).
The KL regularization weight $\beta$ is set to 0.04. The generation parameter is with a sampling temperature as 1, top p as 1, and top k as 0 for diverse generation results in GRPO.
In this stage, the vision encoder is frozen, and the LLM is fine-tuned using Low-Rank Adaptation (LoRA)~\cite{hu2022lora} to reduce the training cost. The LoRA rank is set as 32. This RFT is implemented using VERL training framework.

\begin{table}[H]\footnotesize
\centering
\setlength{\tabcolsep}{6pt}
\renewcommand{\arraystretch}{1.2}
\begin{tabular}{lcccccc}
\toprule
Model & Layers & Hidden size &  Num Heads & Patch Size \\
\midrule
Qwen2.5-VL & 32 & 1280 & 16 & 14 \\

\bottomrule
\end{tabular}
\caption{Model parameters of image encoder model (Vision Transformer).}
\label{tab:table7}
\end{table}

\begin{table}[H]\footnotesize
\centering
\setlength{\tabcolsep}{6pt}
\renewcommand{\arraystretch}{1.2}
\begin{tabular}{lcccccc}
\toprule
Model & Layers & Hidden size & KV Heads & Head Size \\
\midrule

Qwen2.5-VL & 36 & 2048 & 2 & 128 \\
\bottomrule
\end{tabular}
\caption{Model parameters of LLM.}
\label{tab:table7}
\end{table}

\section{Visualization of Dreamed Images}
We selected two representative cases to illustrate the visualized outputs of MindDriver’s predicted future scenes. As depicted in \cref{figdingxing}, the progressive reasoning process is clearly articulated in both examples.

In the first case, MindDriver first performs textual reasoning on the current scene. It identifies pedestrian motion trends, analyzes potential safety risks, and accordingly proposes future behavioral recommendations. Subsequently, based on the outcomes of this textual reasoning, the system generates a visual imagination of the future scene. In the resulting visualization, pedestrian positions can be observed to have changed, demonstrating that our method effectively establishes modeling capabilities for future spatiotemporal relationships. In the second case, a black SUV is crossing the road. If the ego vehicle maintains its current speed, a collision risk exists. Our method accurately imagines the motion state of the vehicle — the black SUV is predicted to reach the center of the road after 0.5 seconds, thereby validating the effectiveness of our approach in both textual reasoning and future imagination.

\section{More Visualization}
We assess the model using closed-loop testing in the CARLA simulator. The model takes visual input in the form of four RGB images from the front-facing camera, encompassing a history of the past two seconds. MindDriver outputs a predicted two-second trajectory, which is then utilized by a PID controller to determine the control signals (throttle, brake, and steering) applied to the vehicle.

\begin{figure*}[t]
    \centering
    \includegraphics[width=0.95\textwidth]{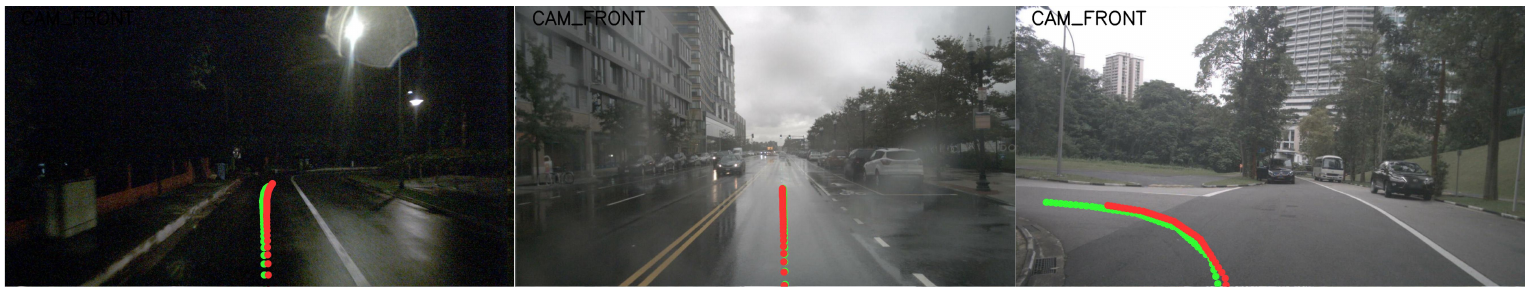} 
    \caption{nuScenes results. The red trajectory is the prediction and the green one is the GT.}
    \label{111} 
\end{figure*}

\textbf{nuScenes results.}
In many challenging nuScenes~\cite{caesar2020nuscenes} scenarios, such as nighttime, heavy rain, and high-curvature roads, MindDriver performs well and avoids collisions. As shown in Fig.~\ref{111}.
In the visualized comparisons, the green trajectory serves as the Ground Truth (GT), while the red trajectory illustrates the path planning executed by MindDriver.
The results exemplify the model's resilience across a spectrum of real-world complexities. In the nighttime scenario (left), despite severe illumination changes and glare, the model maintains a steady path. Similarly, in the overcast and wet urban environment (center), MindDriver adeptly navigates through dynamic traffic, unaffected by the visual noise caused by rain.
Most critically, the turning scenario (right) demonstrates the advantage of our dynamics-driven approach. While traditional methods often struggle with the kinematic constraints of sharp turns, our model produces a smooth trajectory that nearly perfectly overlaps with the GT. This confirms that incorporating dynamics-related rewards significantly enhances the model's ability to handle complex geometric maneuvers with expert-level precision.

\begin{figure*}[t]
    \centering
    \includegraphics[width=0.95\textwidth]{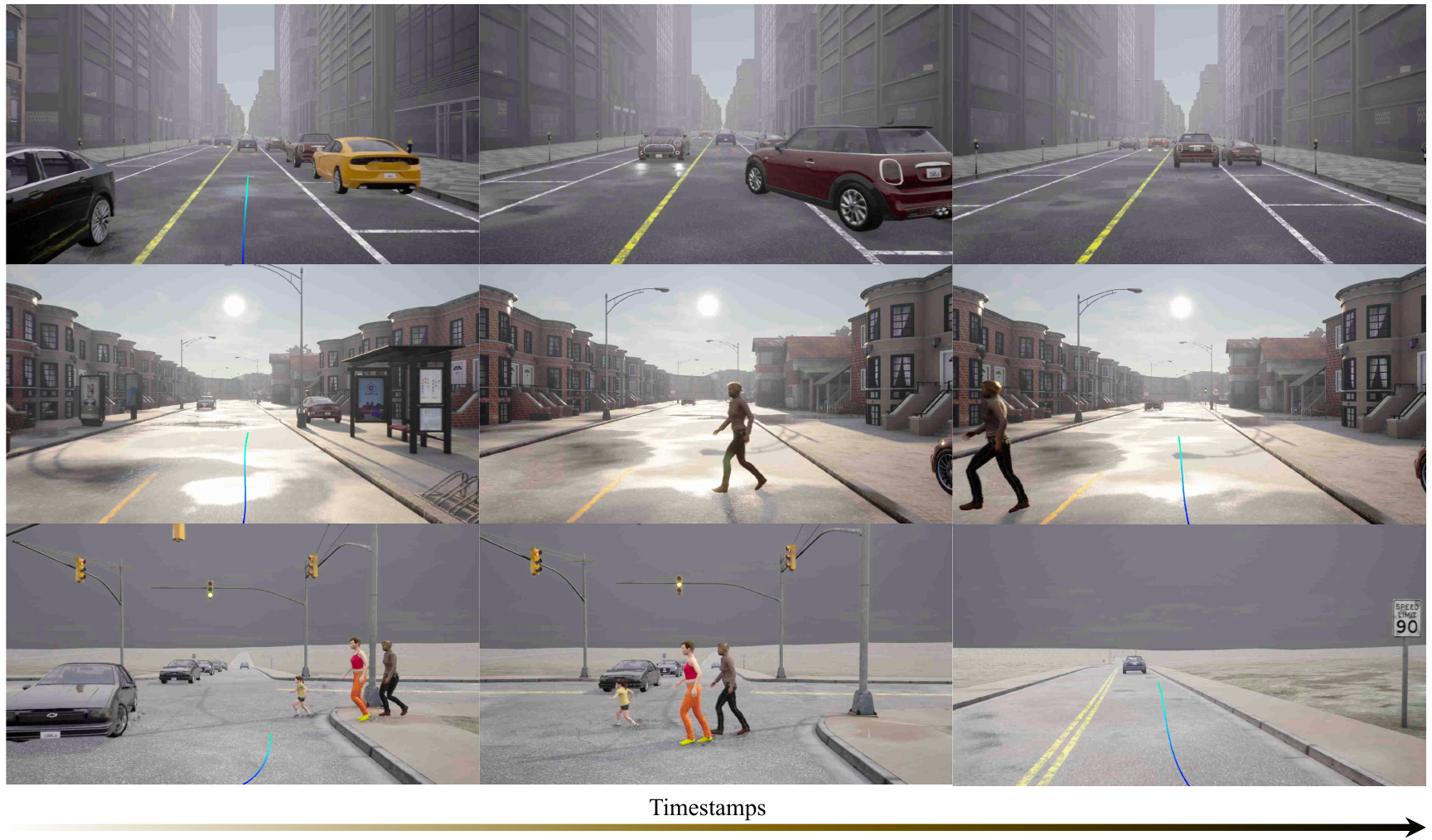} 
    \caption{Bench2drive results. From left to right indicates increasing timestamps.}
    \label{112} 
\end{figure*}

\textbf{Bench2drive results.}
On the simulation dataset there are many scenarios that require risk reasoning. For example, pedestrians crossing the road, narrow roads, nighttime, and other extreme conditions, MindDriver successfully handles them all. As shown in Fig.~\ref{112}.
As illustrated in the visualization, our extensive closed-loop testing on the simulation dataset exposes the model to highly demanding scenarios. The dataset incorporates significant out-of-distribution (OOD) data, ranging from adverse weather conditions with wet road reflections (Top Row) to intense lighting variations.
The model demonstrates remarkable robustness in safety-critical situations. For instance, it effectively anticipates and reacts to dynamic agents, such as vehicles cutting in and pedestrians jaywalking across the street (Middle Row). Notably, in complex intersection scenarios (Bottom Row), the model successfully obeys traffic rules—identifying traffic lights and STOP signs—while making socially compliant decisions to stop and wait for multiple pedestrians, including children. This behavior highlights a significant improvement in planning logic and safety compared to previous SOTA methods like VAD.

{
    \small
    \bibliographystyle{ieeenat_fullname}
    \bibliography{main}
}

\end{document}